\documentclass[peerreview]{IEEEtran}
\usepackage{cite} 
\usepackage{url} 
\usepackage{booktabs} 
\usepackage{graphicx}
\usepackage{multirow}
\usepackage{amsmath}

\begin{document}
\title{Clinnova Federated Learning Proof of Concept: \\ Key Takeaways from a Cross-border Collaboration}


\author{Julia Alekseenko{$^{1,2}$}, Bram Stieltjes{$^{3}$}, Michael Bach{$^{4,5}$}, \\ Melanie Boerries{$^{6, 7}$}, Oliver Opitz{$^{8, 9}$}, Alexandros Karargyris{$^{1}$}, Nicolas Padoy{$^{1,2}$}
\\
$^{1}$ IHU Strasbourg, Institute of Image-Guided Surgery, Strasbourg, France \\
$^{2}$ University of Strasbourg, CNRS, INSERM, ICube, UMR7357, Strasbourg, France
$^{3}$ Research and Analytic Services University Hospital Basel, Basel, Switzerland
$^{4}$ Eye Center, Medical Center – University of Freiburg, Freiburg
Germany
$^{5}$ Faculty of Medicine, University of Freiburg, Freiburg, Germany
$^{6}$ Institute of Medical Bioinformatics and Systems Medicine, Medical Center, University of Freiburg, Faculty of Medicine, University of Freiburg, Freiburg, Germany
$^{7}$ German Cancer Consortium (DKTK) Partner Site Freiburg, a partnership between German Cancer Research Center (DKFZ) and Medical Center, University of Freiburg, Freiburg, Germany
\\ $^{8}$ Coordinating Unit for Digital Medicine Baden-Württemberg (KTBW), Medical Faculty Mannheim, Heidelberg University,
Mannheim, Germany
$^{9}$ Bosch Digital Innovation Hub I Koordinierungsstelle Telemedizin Baden-Württemberg (BDIH I KTBW), Bosch Health Campus (BHC), Stuttgart, Germany
}
\date{25/4/15}

\maketitle
\tableofcontents

\IEEEpeerreviewmaketitle
\begin{abstract}
Clinnova, a collaborative initiative involving France, Germany, Switzerland, and Luxembourg, is dedicated to unlocking the power of precision medicine through data federation, standardization, and interoperability. This European Greater Region initiative seeks to create an interoperable European standard using artificial intelligence (AI) and data science to enhance healthcare outcomes and efficiency. Key components include multidisciplinary research centers, a federated biobanking strategy, a digital health innovation platform, and a federated AI strategy. It targets inflammatory bowel disease, rheumatoid diseases, and multiple sclerosis (MS), emphasizing data quality to develop AI algorithms for personalized treatment and translational research. 

The IHU Strasbourg (Institute of Minimal-invasive Surgery) has the lead in this initiative to develop the federated learning (FL) proof of concept (POC) that will serve as a foundation for advancing AI in healthcare. At its core, Clinnova-MS aims to enhance MS patient care by using FL to develop more accurate models that detect disease progression, guide interventions, and validate digital biomarkers across multiple sites. This technical report presents insights and key takeaways from the first cross-border federated POC on MS segmentation of MRI images within the Clinnova framework. While our work marks a significant milestone in advancing MS segmentation through cross-border collaboration, it also underscores the importance of addressing technical, logistical, and ethical considerations to realize the full potential of FL in healthcare settings.

\end{abstract}

\section{Introduction}
The European Greater Region initiative, involving Grand Est (France), Saarland and Baden-Württemberg (Germany), Basel (Switzerland), and Luxembourg, seeks to develop a sovereign, open, and interoperable European standard to harness AI and data science in healthcare. In particular, this initiative focuses on improving patient outcomes and healthcare efficiency through personalized, sustainable, and preventive solutions. By prioritizing data quality and uniformity, the initiative strives to develop AI algorithms capable of guiding personalized treatment decisions and accelerating translational research into disease etiology, ultimately enhancing patient care. \textit{Central to this initiative is the integration of Federated Learning (FL), a groundbreaking approach to AI model training}. Through the use of FL technology, Clinnova intends to analyze data from various sources, enabling the efficient training of AI algorithms.

Unlike traditional machine learning, which centralizes data for model training, FL enables model training across decentralized clients or centers holding local data samples \cite{xu2021federated}. This approach maintains data privacy by ensuring sensitive information remains within its respective local environment, addressing privacy and data ownership concerns.

FL involves several key steps:
\begin{itemize}
    \item Local Model Training: Local models are trained exclusively on each participating site's own data.
    \item Local Model Updates: After local training, only model updates (e.g., weights) are transmitted to a central server.
    \item Aggregation: The central server aggregates the updates to create a new global model using federated averaging. This updated model is then sent back to each site to start a new round of local training.
\end{itemize}

The Clinnova collaboration exploits this privacy-preserving concept of training to implement, among others, the Clinnova-MS study. This study aims to establish a prospective cohort of multiple sclerosis (MS) patients to study this disorder. Within this study, FL facilitates collaborative AI model training across decentralized data sources, enhancing the ability to develop personalized treatment strategies, identify early MS indicators, and validate digital biomarkers.

This report emphasizes the FL aspect of the Clinnova initiative, detailing the development of two proofs of concept (POCs) as an effort to implement the cross-border collaboration. The first proof of concept (POC1) focuses on deploying a functional virtual infrastructure for FL across multiple centers (clients) and tests this infrastructure using public data. The second proof of concept (POC2) applies this infrastructure to an MS use case using a real, geographically-distributed implementation, a key research focus of Clinnova. The report also addresses practical technical solutions and crucial considerations raised during our work on the POCs.

\section{Motivation}

As part of the cross-border collaborative effort to create the European Valley of Data and AI for Health, the following elements will be developed in Clinnova:

\begin{itemize}
    \item A network of multidisciplinary translational research centers of excellence focused on autoimmune, inflammatory and cancer diseases. 
    \item A federated biobanking strategy allowing to achieve unsurpassed homogeneity for the processing of biological samples collected in the different regions. 
    \item A platform for innovation in digital health based on interoperable data infrastructures established in the different regions.  
    \item \textit{A federated artificial intelligence strategy to enable algorithm training on a massive amount of learning data that are stored in compliance with a shared ethical and regulatory framework.}
\end{itemize}

The IHU focuses on the FL aspect as a key direction of this initiative and leads the implementation of the FL solution. In particular, the objective is to initiate the cross-border European collaboration between Strasbourg, Basel and Freiburg towards running the first FL experiments in the typical scenario of data heterogeneity among centers. In the end, the successful solution permits privacy-preserving collaboration, and can be used to train/validate models without revealing patient data. Thus, resolving issues with data ownership, privacy, and legality.

The primary objective of the Clinnova's POCs is to establish the feasibility of building an FL infrastructure for the experiments, promote collaboration between participating centers, and investigate the potential power, needs, and challenges of its development, rather than proposing a novel federated deep learning solution.

To achieve the Clinnova federated AI strategy, the development of POC 1 and POC 2 is an initial step. To be more precise, POC 1 aims at the feasibility of performing FL between clients for the overall technical goal of Clinnova. Specifically, it targets the following: 

\begin{itemize}
\item Deploying a virtual FL infrastructure across the centers (clients).
\item Starting, running, and stopping training/validation experiments on the FL infrastructure.
\item Running trial experiments using a public multi-center medical imaging dataset.
\item Identifying issues with FL deployment and continuous FL workflow execution (e.g., network issues, up/downtime issues).
\item Functioning as the precursor to the Clinnova FL workflows.
\end{itemize}
\
While POC 2 extends these targets to: 
\begin{itemize}
\item Deploying a real (geographically-distributed) FL infrastructure and extending the network of collaborative centers.
\item Applying the infrastructure to the MS use case, aligning with Clinnova's scope of interest.
\item Identifying issues with FL workflow execution under this distributed setup.
\end{itemize}

\section{FL System Elements}
\subsection{Platform}
sciCORE+ \cite{scicore_website} is a secure research platform that offers a robust computing environment, empowering users to seamlessly transfer, store, manage, and analyze sensitive research data.

sciCORE+ offers a scalable OpenStack infrastructure, combining the computing power of a midsize high-performance computing (HPC) system. Equipped with multi-CPU and multi-GPU compute nodes, it provides a versatile environment to accommodate diverse computational workloads. Moreover, sciCORE+ prioritizes data security by offering support for data encryption, secure backup solutions, and private-cloud environments.

The SciCOREmed environment is utilized as the pre-made solution offered by University of Basel, Switzerland. It has straightforward prerequisites for parties to join it, including Terms of Use, 2FA-SSH (maintenance access), HTTPS, and other protocols (service interoperability).

\subsection{FL Framework} We employ NVIDIA FLARE \cite{roth2022nvidia} as our FL framework due to its established familiarity among several participating centers involved in the POC development. This pre-existing familiarity with FLARE significantly accelerated the development process, as it streamlined implementation and reduced the learning curve for the involved teams. FLARE’s robust features, including its support for various machine learning and deep learning pipelines, make it exceptionally well-suited for our requirements. Its flexibility and scalability enable seamless integration with diverse data sources and facilitate the development of complex models. Additionally, FLARE’s strong community support and comprehensive documentation further enhance its effectiveness.

\subsection{Important Considerations}

\textbf{Future Challenges:} Although sciCORE+ is a viable solution for the POCs, there might be a need to develop a new platform in the future. This could be driven by several factors, such as its inability to adapt to all the requirements that may arise as the project progresses. Indeed, while for POC 1, the platform fully supports the entire infrastructure by simulating a real FL scenario, for POC 2 this platform is used only for hosting an aggregator to provide a secure connection and aggregation; the question of where and how to better host it could arise as well.

\textbf{Onboarding:} The introduction of new technologies to users may demand sufficient training and support to ensure seamless adoption. Users might encounter difficulties in navigating the platform, efficiently managing resources, or resolving technical issues without appropriate training and guidance.  

\textbf{Team FL Expertise:} Effective experiment management in FL relies heavily on center personnel possessing a strong understanding of FL concepts. Ideally, this should be coupled with familiarity with a specific FL framework. It is important to acknowledge that there might be disagreements regarding the preferred FL framework. Reaching a consensus on a suitable framework (or support of multiple options) is essential. 

\section{Proof of Concept 1}
\begin{table*}[!t]
\centering
\caption{Prostate Segmentation Dataset Used in POC 1}
\label{tab:Prostate Dataset}
\resizebox{0.9\textwidth}{!}{%
\begin{tabular}{c c c c c c c}
\hline
\textbf{Origin} & \textbf{Client-tenant} & \textbf{Scanner} & \textbf{Field} & \textbf{Resolution (mm$^3$)} & \textbf{Train} & \textbf{Val} \\
\hline
I2CVB\cite{lemaitre2015computer}  & Strasbourg & Siemens, GE & 3.0, 1.5 T & 0.67-0.79/1.25, 0.27x0.27/3.00-3.50 & 32 & 7 \\ 
\hline
MSD\cite{simpson2019large} & Basel & Siemens & 3.0 T & 0.60-0.625/3.60-4.00 & 26 & 6 \\
\hline
NCI-ISBI\cite{bloch2015nci}&Mock & Philips & 1.5 T & 0.40x0.40x3.00 & 32 & 7 \\
\hline
\end{tabular}%
}
\end{table*}

\subsection{Virtual Infrastructure}
sciCORE+ provides tenants for each client with virtual services for computing, networking, and storage capacity. The decision to utilize virtual services for POC 1 is based on several factors:

\begin{itemize}
    \item Rapid Deployment: Virtual services facilitate fast provisioning and deployment, allowing for the rapid setup and configuration of the initial prototype. This expedites the development cycle, enabling quicker iterations and testing of diverse features and functionalities.
    
    \item Isolation: Each tenant receives its dedicated virtual environment, ensuring robust isolation and security among different clients. This setup closely mirrors the real-world FL environment, where data privacy and security are paramount.
    
    \item Ease of Management: Virtual environments offer streamlined management and maintenance processes compared to physical hardware. Consequently, the FL IT administrator has full access rights to each tenant, enabling seamless connection/disconnection, updates, and necessary maintenance steps for FL experiments.
\end{itemize}

Each tenant has its own Titan XPR GPU, CPU, and storage space. At this stage of the project only Basel and Strasbourg are involved in the project, with an additional mock-tenant included to enhance the FL network: 

\begin{itemize}
    \item \textbf{Strasbourg-tenant:}
    \begin{itemize}
        \item Dataset 1
        \item CPU capacity 1
        \item GPU capacity 1
        \item Storage Space 1
    \end{itemize}
    \item \textbf{Basel-tenant:}
    \begin{itemize}
        \item Dataset 2
        \item CPU capacity 2
        \item GPU capacity 2
        \item Storage Space 2
    \end{itemize}
    \item \textbf{Mock-tenant:}
    \begin{itemize}
        \item Dataset 3
        \item CPU capacity 3
        \item GPU capacity 3
        \item Storage Space 3
    \end{itemize}
\end{itemize}

All libraries, frameworks, and necessary dependencies are integrated into a Docker container and uploaded to DockerHub.

\subsection{Prostate Segmentation Use Case}
To validate the feasibility of our developed infrastructure for multi-centric research, we aim to conduct experiments using real-world multi-centric public data in an FL fashion. Given the availability of such data, we use a widely recognized prostate dataset (detailed later in \ref{prostate_data}) as an ideal candidate for our test experiments and an important use case.

Prostate cancer (CaP) is a significant global health issue, and accurate segmentation of prostate images using magnetic resonance imaging (MRI) is essential for diagnosing and localizing this disease. According to Ferlay et al. \cite{ferlay2015cancer}, prostate cancer (CaP) is a major concern, ranking among the top five cancers in terms of both incidence and mortality. It is the most commonly diagnosed cancer in men, with approximately 1.6 million new cases reported globally in 2015 \cite{global2017global}. Magnetic resonance imaging (MRI) plays a crucial role in diagnosing and localizing CaP.

\subsection{Dataset}\label{prostate_data}
To perform experiments as part of POC 1, we utilize a public multi-site dataset for prostate MRI segmentation \cite{lemaitre2015computer}, \cite{simpson2019large}, \cite{bloch2015nci}. This dataset includes prostate T2-weighted MRI images and corresponding segmentation masks.  The split between three clients (tenants) and public data is presented in Table \ref{tab:Prostate Dataset}.

In the data preparation phase, each client starts by downloading the dataset from the source. The dataset is then cleaned to remove problematic cases, such as missing slices or mismatched images and labels. Following this, thresholding is applied to combine two label values into a single mask of the prostate.

The preprocessing phase involves a series of normalization steps. First, voxel normalization is performed to rescale MRIs to a resolution of 0.30x0.30x1.00 (mm$^3$). The orientation of the input image is then adjusted to the “RAS” (Right-Anterior-Superior) axis codes. Finally, Z-score normalization is applied to standardize the intensity values across the dataset.

\subsection{Implementation Details}
\textbf{Network Architectures:} We use U-Net \cite{ronneberger2015u} as our segmentation baseline. The accuracy of the segmentation model is measured using the average Dice score.

\textbf{Federated Algorithms:} The following algorithms are evaluated:
\begin{itemize}
    \item FedAvg: The Federated Averaging (FedAvg) algorithm conducts multiple local stochastic gradient updates at client nodes and averaging of the model parameters at the server \cite{mcmahan2017communication}.
    \item FedProx: This is a generalization of FedAvg that incorporates modifications to better handle data and system heterogeneity. Clients in FedProx optimize a regularized loss function that includes a proximal term \cite{li2020federated}.
    \item Ditto: The algorithm focuses on learning personalized models for each client. When the regularization hyper-parameter is zero, Ditto reduces to learning separate local models, whereas FedProx would reduce to FedAvg \cite{li2021ditto}.
\end{itemize}

\section{Proof of Concept 2}

\subsection{Distributed Infrastructure}

\begin{figure*}[ht]
    \centering
    \includegraphics[width=0.80\textwidth]{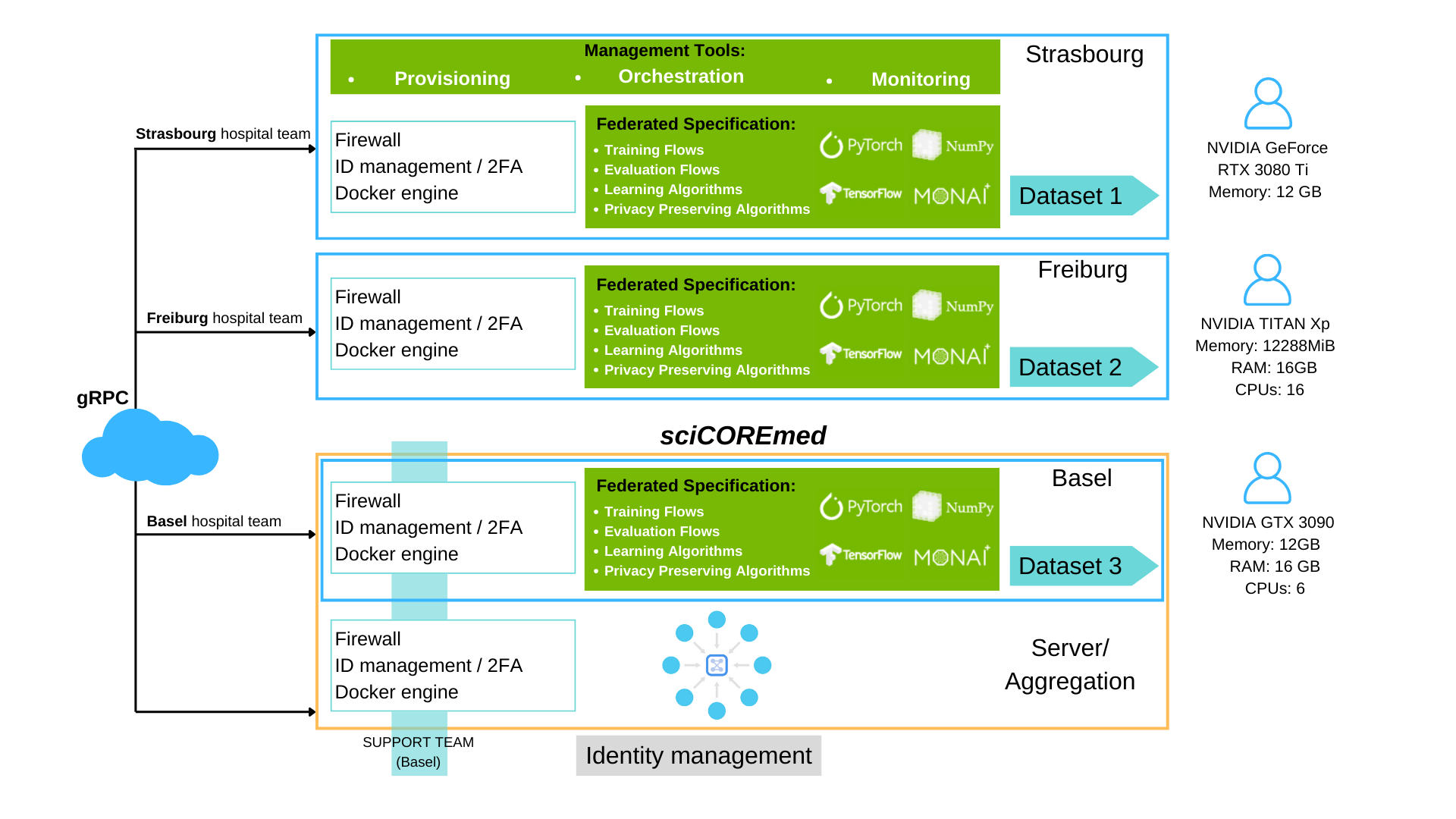}
    \caption{Real-world Geographically-distributed FL Infrastructure for POC 2}
    \label{fig:platform}
\end{figure*}

For POC 2, the FL infrastructure is extended to a real-world geographically-distributed scenario, ensuring secure data handling and efficient computing infrastructure across clients (as shown in Figure \ref{fig:platform}). This expansion included:

\begin{itemize}
    \item New Client Participation: Freiburg has joined the project, eliminating the need for a mock tenant as in POC 1.
    
    \item Decentralized Control: In contrast to POC 1, each center in the extended POC 2 retains autonomy and control over its resources.
    
    \item Resource Allocation: Each center is equipped with its own resources, such as GPUs, CPU, and storage space.
\end{itemize}

This environment provides a more realistic representation of real-world setups, enabling a more accurate assessment and validation of proposed solutions.

\subsection{Multiple Sclerosis Use Case}

Multiple sclerosis (MS) is a chronic autoimmune disease of the central nervous system (CNS) characterized by inflammation, demyelination, and neurodegeneration. In MS, the immune system mistakenly attacks the myelin sheath, a fatty substance that surrounds and insulates nerve fibers in the brain, spinal cord, and optic nerves \cite{matute2018multiple}. 

Despite significant advancements in MS research, many aspects of the disease remain poorly understood, including its etiology, heterogeneity, progression, and optimal treatment strategies. Challenges persist in accurately diagnosing MS, predicting disease course, and developing targeted therapies that effectively halt or reverse disease progression without significant side effects. Furthermore, there is a need for biomarkers that can reliably monitor disease activity and response to treatment, as well as tools for personalized medicine approaches tailored to individual patient characteristics \cite{gholamzad2019comprehensive}.

For POC 2, we strategically select a publicly available dataset that directly aligns with Clinnova's core objective of tackling MS. This dataset, the Shift Benchmark on Multiple Sclerosis lesion segmentation data \cite{malinin2021shifts}, \cite{malinin2022shifts}, provides an ideal case study to evaluate the effectiveness of FL models within our research framework.

The task at hand involves segmenting MS in MRIs, a critical aspect for understanding disease progression and treatment effectiveness. Accurate MS segmentation enables precise analysis of disease burden, assisting clinicians in treatment planning and monitoring. Given the variability in MS characteristics such as size, shape, achieving accurate segmentation is essential for comprehensive disease management.

However, manual segmentation remains challenging and time-consuming, even for experienced clinicians, due to physiological variations in tissue intensities, shapes, and sizes, resulting in significant inter-subject variability. Furthermore, challenges are exacerbated by variations in imaging protocols, the use of coils, and demographic differences among data from multiple medical centers.

Recent advancements in deep neural networks (DNNs) have enabled automatic segmentation of multiple sclerosis (MS) lesions \cite{shoeibi2021applications}. By leveraging multiple data sources for training robust and accurate segmentation DNNs, we can potentially develop a more generalized model. In this context, the FL infrastructure becomes particularly valuable, as it facilitates multi-centric research and address the complexities associated with heterogeneous medical data.

\subsection{Dataset}
The Shift Benchmark on Multiple Sclerosis lesion segmentation dataset provides the following characteristics:

\begin{table*}[!t]
\centering
\caption{Multiple Sclerosis Lesion Segmentation Dataset Used in POC 2}
\label{tab:MS Dataset}
\resizebox{0.8\textwidth}{!}{%
\begin{tabular}{c c c c c c c c}
\hline
\textbf{Origin} & \textbf{Client} & \textbf{Scanner} &  \textbf{Field}  & \textbf{Resolution (mm$^3$)} & \textbf{Raters}  & \textbf{Train}  & \textbf{Val} \\
\hline
Rennes &                        & S Verio & 3.0 T & 0.50×0.50×1.10 & 7 &  - & -  \\
Bordeaux & Basel & GE Disc  & 3.0 T & 0.47×0.47×0.90 & 7  &  11 & 12  \\
                         & Freiburg & S Aera  & 1.5 T & 1.03×1.03×1.25 & 12  &  12 & 12  \\
                   Lyon &                        & P Ingenia  & 3.0 T & 0.74×0.74×0.70 & 7  &  - & -  \\
                   \hline
                  Brest & Strasbourg  & P Medical  & 3.0 T & 0.82×0.82×2.20 & 2  &  10 & 9  \\
\hline
\end{tabular}%
}
\end{table*}

\begin{itemize}
    \item Multi-center: Data collected from multiple medical centers (Rennes, Bordeaux, Lyon, Brest), providing diverse perspectives and insights into lesion segmentation.
    \item Multi-modality: Incorporates data from various imaging modalities such as FLAIR and T1, enriching the dataset with different imaging perspectives and enhancing segmentation accuracy. 
    \item Public: Accessible to the research community, encouraging collaboration, transparency, and advancement in segmentation research.
\end{itemize}
 
Limitations of the dataset include: 

\begin{itemize}
    \item Preprocessing: Data has undergone preprocessing steps, potentially limiting the flexibility for researchers to apply their own preprocessing techniques on raw MRIs.
    \item Lack of metadata (i.e., center splits): As data from Rennes, Bordeaux, and Lyon are provided within a single folder without specifying the origin of each data point, center-specific segmentation performance analysis is hindered due to the absence of metadata.
\end{itemize}

We use 3D brain MRIs acquired using both FLAIR and T1 contrasts. Patient MRIs from different locations exhibit variations in terms of scanner models, local annotation (rater) guidelines, scanner strengths (1.5T vs 3T), and resolution of the raw MRIs. The split between three clients and public data is presented in Table \ref{tab:MS Dataset}. Specifically, Strasbourg receives data exclusively from Brest, while the remaining datasets are distributed between Basel and Freiburg.

Each sample undergoes preprocessing, which includes denoising, skull stripping (where the brain mask is learned from the T1 modality image registered to the FLAIR space), bias field correction, and interpolation to a 1 mm$^3$ isovoxel space. The ground-truth segmentation mask is also interpolated to the 1 mm$^3$ isovoxel space.

\subsection{Implementation Details}

\textbf{Network Architectures:} Our FL implementation utilizes well-established segmentation models from the literature: U-Net \cite{ronneberger2015u}, SegResNet \cite{myronenko20193d}, Attention U-Net \cite{oktay2018attention}, Swin UNETR \cite{hatamizadeh2021swin}, and UNETR \cite{hatamizadeh2022unetr}. The accuracy of the segmentation models is measured using the average Dice score.

\textbf{Federated Algorithm:} FedAvg \cite{mcmahan2017communication} is an optimization strategy for MS experiments. While we recognize its limitations, particularly with strong non-IID data distributions, this choice remains optimal for POC 2 to maintain a reasonable number of experiments. Also, given that Basel and Freiburg share data from the same pool, we believe the non-IID problems are not significant.

\section{Collaborative Network}
\subsection{Development of the Collaboration}
\textit{Participating Centers:}
The Clinnova POCs involve three healthcare institutions located in France (Strasbourg), Germany (Freiburg), and Switzerland (Basel).

Processes: Regular communication is maintained through monthly video conferences and a shared project collaborative platform. Roles and responsibilities are clearly defined in a project charter, outlining each center's obligations and contributions. Each center appoints a center coordinator who ensures timely communication and execution of tasks.

\textit{Degree of Familiarity or Trust Between Participating Centers:}
Some centers have pre-existing partnerships and trust (e.g., between France and Switzerland after POC 1), while others are new collaborators (Freiburg in POC 2).

Processes: Independent facilitators mediate discussions to align on shared objectives. Regular progress reviews and feedback sessions ensure transparency and build trust among participants.

\textit{Knowledge Sharing and Capacity Building:}
The initiative promotes continuous learning and improvement among its members in terms of FL.

Processes: Joint training sessions, workshops, or knowledge repositories to share best practices in FL are organized. These sessions are tailored to address specific challenges faced by each center.

\textit{Data Security and Privacy Compliance:}
Ensuring data security and privacy is essential in Clinnova, given the sensitive nature of healthcare data.

Processes: Each participating center complies with national and international regulations, such as GDPR in Europe. 

\textit{Evaluation and Monitoring:}
Continuous evaluation and monitoring are essential to measure the success and impact of the POC development.

Processes: Key performance indicators (KPIs) are defined to track progress and outcomes. These include metrics related to project milestones, participant engagement.

\subsection{Important Considerations}

\textbf{Communication and Coordination:}
    \begin{itemize}
        \item Scheduling Conflicts: Monthly video conferences can lead to scheduling conflicts, making it difficult for all participants to attend regularly.
    \end{itemize} 

\textbf{Trust and Collaboration Among Partners:}
\begin{itemize}
    \item  Building Trust: Establishing trust with new centers can be time-consuming, particularly when there is no prior working relationship.
    \item Alignment of Objectives: Differing priorities and objectives among partners can create misalignment.
    \item Resource Allocation: Uneven distribution of resources, such as funding, technology, and personnel, can create disparities in contributions and expectations, leading to tension and dissatisfaction among centers.
\end{itemize}
   
\textbf{Knowledge Sharing and Capacity Building:}
 
\begin{itemize}
    \item Resistance to Change: Resistance to adopting new technologies or methodologies for FL may hinder capacity building efforts, especially among centers with established practices.
\end{itemize}

\textbf{Technical and Logistical:}
\begin{itemize}
    \item Infrastructure Disparities: Differences in technological infrastructure and computational resources across centers can affect the uniformity.
    \item Deployment of Federated Learning: Ensuring efficient and secure deployment of FL in a geographically-distributed setup requires significant coordination and technical expertise.
\end{itemize}

\textbf{Regulatory and Compliance:}
\begin{itemize}
    \item Varying Regulations: Navigating the different data protection regulations and ethical guidelines across countries can be complicated, potentially leading to compliance issues.
\end{itemize}

\textbf{Evaluation and Monitoring:}
\begin{itemize}
    \item Resource Constraints: Limited resources, including time, personnel, and budget, may constrain the ability to conduct thorough evaluation and monitoring activities, potentially leading to incomplete or biased assessments.
\end{itemize}
    
\section{Evaluations and Experiments}
\subsection{Proof of Concept 1}

\textbf{Client Resource Allocation Experiment:} 

The crucial part of successfully running the experiments is providing efficient computational resources. While storage is typically less critical since sites have their own infrastructure, difficulties can arise with computational power. In this set of experiments presented in Table \ref{tab:hardware}, we focus on testing various setups for allocating computational resources to determine how these allocations affect the overall performance of the FL federation.

\begin{table}[ht]
\centering
\caption{Hardware Configuration for FL Experiments in POC 1}
\resizebox{0.47\textwidth}{!}{%
\begin{tabular}{ccccc}
\hline
\textbf{Experiment} & \textbf{Aggregator} & \textbf{Strasbourg} & \textbf{Basel} & \textbf{Mock}  \\
\hline
1: all CPUs &  & CPU & CPU & CPU  \\
2: one GPU & FedAvg & GPU & CPU & CPU \\
3: all GPUs &  & GPU & GPU & GPU  \\
\hline
\end{tabular}
}%
\label{tab:hardware}
\end{table}

While aggregation and experiment fetching (less than 1\% of the total time) are not the most time-consuming tasks, training and waiting times account for the largest share of the total duration. Additionally, these steps are not highly computationally intensive, so they could potentially be performed without GPU support if server resources are limited. 

A brief analysis of the results presented in Table \ref{tab:time} reveals that the experiment with CPU clients took the longest, while Experiment 2 was the second slowest. The fastest results were achieved when all clients were equipped with GPUs. Even with GPUs that were not the most powerful, we observed a 32.14\% improvement compared to Experiment 1. Although this outcome is expected, it is noteworthy that, with a small dataset, the time differences between experiments are substantial. This underscores the importance of using powerful GPUs on each client for large-scale experiments with extensive datasets within the Clinnova project.

\begin{table}[ht]
\centering
\caption{Average Total Time in Hours (hr) Spent on Training, Aggregation and Validation in POC 1}
\resizebox{0.5\textwidth}{!}{%
\begin{tabular}{ccccc}
\hline
\textbf{Experiment} & \textbf{Train+Validate$\approx$} & \textbf{Aggregate$\approx$} & \textbf{Train$\approx$} & \textbf{Validate$\approx$} \\
\hline
1: all CPUs & 64.18 hr & & 63.74 hr & 0.44 hr \\ 
2: one GPU & 50.33 hr & 0.09 hr & 50.04 hr & 0.29 hr \\ 
3: all GPUs & 43.55 hr & & 43.36 hr & 0.19 hr \\
\hline
\end{tabular}
}%
\label{tab:time}
\end{table}

\begin{table}[]
\centering
\caption{Average Round and Waiting Time in Hours (hr) in POC 1}
\label{tab:round_time}
\resizebox{0.5\textwidth}{!}{%
\begin{tabular}{lllll}
\hline
 \textbf{Client-Tenant} & \textbf{First Round$\approx$} & \textbf{Last Round$\approx$} & \textbf{Avg Round$\approx$} & \textbf{Waiting$\approx$} \\
 \hline
\multicolumn{5}{l}{Experiment 1: all CPUs} \\
\hline
Strasbourg & 0.59 hr & 0.57 hr & 0.58 hr & 0.01 hr \\ 
Basel & 0.28 hr & 0.40 hr & 0.34 hr & 0.35 hr \\ 
Mock & 0.40 hr & 0.56 hr & 0.48 hr & 0.21 hr \\ 
\hline 
\multicolumn{5}{l}{Experiment 2: one GPU} \\ 
\hline 
Strasbourg & 0.42 hr & 0.41 hr & 0.41 hr & 0.06 hr \\ 
Basel & 0.31 hr & 0.40 hr & 0.35 hr & 0.15 hr \\ 
Mock & 0.48 hr & 0.49 hr & 0.48 hr & 0.01 hr \\ 
\hline 
\multicolumn{5}{l}{Experiment 3: all GPUs} \\ 
\hline 
Strasbourg & 0.42 hr & 0.40 hr & 0.41 hr & 0.01 hr \\ 
Basel & 0.23 hr & 0.30 hr & 0.27 hr & 0.20 hr \\ 
Mock & 0.26 hr & 0.28 hr & 0.27 hr & 0.18 hr \\
\hline
\end{tabular}%
}
\label{tab:time1}
\end{table}

Moreover, Table \ref{tab:time1} demonstrates that a client with slower hardware adversely affects the performance of other clients. For example, even if client A completes the round quickly, the overall progress is delayed by the slowest client’s submission. Although modifying the experiment logic could partially address this issue, careful consideration is needed to ensure efficient operation of the FL federation.

\textbf{Performance Experiment:}

The initial results, as shown in Table \ref{tab:first-results}, are promising. Notably, the highest average Dice score achieved is 0.609. Analyzing the results per client reveals even higher scores, except for Strasbourg, which may be affected by the dataset's complexity, as it comprises two distinct parts \cite{lemaitre2015computer}. FedAvg is the most effective aggregation algorithm, though Ditto also shows potential. Conversely, FedProx underperformed, which is surprising given its purpose of addressing heterogeneity. Fine-tuning the algorithm's parameters might enhance its performance.



\begin{table}[ht]
\centering
\caption{Dice Score ($\pm$ std) of Different Aggregation Algorithms on U-Net Models Across Client-Tenants in POC 1}
\label{tab:first-results}
\resizebox{0.5\textwidth}{!}{%
\begin{tabular}{cccccc}
\hline
\textbf{Model} & \textbf{Aggregation} & \textbf{Strasbourg} & \textbf{Basel} & \textbf{Mock} & \textbf{Global Mean} \\
\hline
\multirow{2}{*}{U-Net} & FedAvg  & 0.413 & 0.636 & 0.779 & \textbf{0.609} \\
& & $\pm$0.20 & $\pm$0.13 & $\pm$0.09 & \textbf{$\pm$0.15} \\
\cline{2-6}
& FedProx & 0.274 & 0.510 & 0.618 & 0.467 \\
& & $\pm$0.24 & $\pm$0.15 & $\pm$0.08 & $\pm$0.14 \\
\cline{2-6}
& DITTO & 0.371 & 0.452 & 0.832 & 0.551 \\
& & $\pm$0.21 & $\pm$0.16 & $\pm$0.07 & $\pm$0.20 \\
\hline
\end{tabular}%
}
\end{table}

\textbf{Infrastructure Aspects:}

When building a virtual FL infrastructure for POC 1, the following aspects were discussed and evaluated:

\begin{itemize}
    \item Resource Allocation: Considering different resource allocation strategies to optimize the utilization of GPUs, CPUs, and storage space among tenants. This could provide practical guidelines for clients joining during POC 2, helping them navigate resource allocation for further experiments.
    
    \item Performance Benchmarking: Benchmarking the performance of the virtual setup under varying workloads to assess its computational capabilities. 
    
    \item Security: Analyzing security levels to identify potential vulnerabilities in the infrastructure and evaluating the effectiveness of existing security measures.
    
    \item Scalability: Assessing the ability of the infrastructure to handle additional computational workloads.
    
    \item Fault Tolerance: Evaluating the fault tolerance capabilities of the infrastructure and assessing its resilience to failures and disruptions.
    
\end{itemize}

The conclusions drawn from these evaluations include:

\begin{itemize}
    \item Optimal resource allocation strategies were identified, leading to improved utilization of computational resources for further experiments in POC 2. As shown in Tables \ref{tab:time} and \ref{tab:time1}, we focused on the computational resources aspect. Before integrating a new client into the federation, we discuss various possibilities, such as whether the client has GPUs or only CPUs, the potential configurations, whether the resources can be fully dedicated to the FL experiment, whether multi-GPU support is available, and the storage requirements. Our best strategy is to maximize computational performance at each site by analyzing the resources and striving to balance them among the clients.
    
    \item The infrastructure demonstrated good performance under varying workloads related to MRI data. Initially, we handled preprocessing tasks, such as organizing and normalizing numerous MRI scans and slices. Following this, we addressed more complex tasks, including segmentation of the MRI data. This range from basic preprocessing to advanced segmentation highlights the infrastructure's capability to efficiently manage both simple and complex computational tasks.
    
    \item Security assessments revealed potential vulnerabilities, such as poisoning or malicious clients, insufficient encryption of data in transit, and inadequate authentication mechanisms. These issues should be taken into consideration when dealing with real private data, enhancing the overall security of the proposed solution.
    
    \item Scalability evaluation showed that the infrastructure could support a high number of tenants, with minimal impact on performance. We replicated multiple instances within each tenant, evaluating the communication speed between the server and tenants, as well as the ease of managing the federation. Both the framework and virtual infrastructure remained stable, with no limitations observed in this regard.
    
    \item Fault tolerance evaluations demonstrated the platform's resilience to some failures and network disruptions. Generally, the platform handled long experiments well, and in the event of failures, the ability to continue the experiment was supported. If a server fails, the clients automatically reconnect until the connection is successful, requiring only the server to be restarted, not each individual client. If a client fails, it needs to re-initiate the connection to the server, while the rest of the federation can either wait for the connection or continue the experiment. By default, the virtual environment is quite manageable in the face of network failures. As such, only some common cases were artificially tested.
\end{itemize}

Overall, the experiments in POC 1 provided valuable insights into the capabilities and limitations of the proposed infrastructure solution.

\subsection{Proof of Concept 2}

\textbf{Performance Experiment:}

\begin{table}
\centering
\caption{Dice Score ($\pm$ std) of Federated Global Models on Segmentation Task: FLAIR in POC 2}
\resizebox{0.47\textwidth}{!}{%
\label{tab:segmentation_performance0}
\begin{tabular}{c c c c c}
\hline
\textbf{Model} & \textbf{Basel} & \textbf{Freiburg} & \textbf{Strasbourg} & \textbf{Global Mean} \\
\hline
U-Net & 0.582 & 0.630 & 0.595 & 0.602 \\
& $\pm$0.23 & $\pm$0.14 & $\pm$0.06 & $\pm$0.02 \\
\hline
SegResNet & 0.625 & 0.648 & 0.601 & \textbf{0.625} \\ & $\pm$0.21 & $\pm$0.11 & $\pm$0.08 & \textbf{$\pm$0.02} \\
\hline
Attention U-Net & 0.602 & 0.671 & 0.586 & 0.619 \\  & $\pm$0.23 & $\pm$0.13 & $\pm$0.09 & $\pm$0.04 \\
\hline
SwinUNETR & 0.605 & 0.662 & 0.591 & 0.619 \\  & $\pm$0.22 & $\pm$0.13 & $\pm$0.09 & $\pm$0.04 \\
\hline
UNETR & 0.579 & 0.618 & 0.568 & 0.588 \\  & $\pm$0.21 & $\pm$0.14 & $\pm$0.08 & $\pm$0.03 \\
\hline
\end{tabular}
}
\end{table}

Based on the provided performance metrics for various segmentation models across different locations for the FLAIR sequence (see results in Table\ref{tab:segmentation_performance0}), we conclude that among the models evaluated (U-Net, SegResNet, Attention U-Net, SwinUNETR, and UNETR), SegResNet consistently achieves the highest Dice Score across all clients, with values ranging from 0.601 to 0.648.

\begin{table}
\centering
\caption{Dice Score ($\pm$ std) of Federated Global Models on Segmentation Task: FLAIR and T1 in POC 2}
\resizebox{0.47\textwidth}{!}{%
\label{tab:segmentation_performance1}
\begin{tabular}{c  c c c  c}
\hline
\textbf{Model} & \textbf{Basel} & \textbf{Freiburg} & \textbf{Strasbourg} & \textbf{Global Mean} \\
\hline
U-Net  & 0.590 & 0.647 & 0.567 & 0.601  \\  & $\pm$0.22 & $\pm$0.12 & $\pm$0.05 &  $\pm$0.04\\
\hline
SegResNet & 0.608 & 0.675 & 0.628 & \textbf{0.637} \\  & $\pm$0.23 & $\pm$0.10 & $\pm$0.07 & \textbf{$\pm$0.03}  \\
\hline
Attention U-Net & 0.609 & 0.673 & 0.593 & 0.625  \\  & $\pm$0.22 & $\pm$0.11 & $\pm$0.07 & $\pm$0.04  \\
\hline
SwinUNETR & 0.610 & 0.686 & 0.611 & 0.636  \\  & $\pm$0.23 & $\pm$0.12 & $\pm$0.08 & $\pm$0.04   \\
\hline
UNETR & 0.602 & 0.661 & 0.573 & 0.612  \\  & $\pm$0.23 & $\pm$0.13 & $\pm$0.06 & $\pm$0.04  \\
\hline
\end{tabular}
}
\end{table}

Based on the provided Dice Scores for FLAIR + T1 lesion segmentation across different clients (see results in Table \ref{tab:segmentation_performance1}), we notice that across all clients, SegResNet also achieves the highest Dice Score, followed closely by SwinUNETR and Attention U-Net. Across all models, the inclusion of T1 contrasts alongside FLAIR generally leads to a slight improvement in Dice Scores. There are minor fluctuations in performance across different clients, but the overall trends are consistent.

\textbf{Global vs Local Experiment:}
\begin{table}[htbp]
\centering
\caption{Dice Score ($\pm$ std) of Federated Global Model and Cross-center Local Training/Validation and Loss  ($\%$) Against Global Model in POC 2}
\resizebox{0.43\textwidth}{!}{%
\label{tab:segmentation_performance2}
\begin{tabular}{c c c c}
\hline
\textbf{Trained / Validated} & \textbf{Basel} & \textbf{Freiburg} & \textbf{Strasbourg} \\
\hline
{Global} & \textbf{0.608} & \textbf{0.675} & \textbf{0.628} \\ & \textbf{$\pm$0.23} & \textbf{$\pm$0.10} & \textbf{$\pm$0.07} \\
\hline
{Basel} & 0.365 & 0.315 & 0.510 \\
 & -24.30\% & -36.00\% & -11.80\% \\
\hline
{Freiburg} & 0.570 & 0.633 & 0.565 \\
 & -3.80\% & -4.23\% & -6.30\% \\
\hline
{Strasbourg} & 0.556 & 0.619 & 0.615 \\
 & -5.20\% & -5.60\% & -1.30\% \\
\hline
\end{tabular}
}
\end{table}

From the provided experiments comparing the SegResNet model trained and evaluated globally with FLAIR + T1 data against locally trained and evaluated models in Basel, Freiburg, and Strasbourg (see the results in Table \ref{tab:segmentation_performance2}), we conclude the following:

\begin{itemize}
    \item Dice Score: The SegResNet model trained and evaluated globally achieves higher Dice Scores compared to locally trained models across all three clients. For example, in Basel, the global SegResNet achieves a Dice Score of 0.608, while the locally trained model achieves only 0.365. Similar trends are observed in Freiburg and Strasbourg.
    
    \item Consistency Across Locations: Despite variations in performance across different locations, the global SegResNet consistently outperforms locally trained models in terms of Dice Scores. This indicates the robustness and generalizability of the globally trained model across diverse datasets and settings.
\end{itemize}

\textbf{Client Data Quantity Experiment:}

\begin{table}[htbp]
\centering
\caption{Dice Score ($\pm$ std) of Federated Global Model and Locally Trained on Segmentation Task under Different Basel Client Data Quantity in POC 2}
\resizebox{0.38\textwidth}{!}{%
\label{tab:segmentation_performance3}
\begin{tabular}{c c c c}
\hline
\textbf{Global} & \textbf{Local} & \textbf{Global} & \textbf{Global} \\
\textbf{Basel} & \textbf{Basel} & \textbf{Freiburg} & \textbf{Strasbourg} \\
\hline
\multicolumn{4}{c}{\text{20\% of Data on Basel Client:}} \\
\hline
\textbf{0.598} & 0.475 & 0.655 & 0.603 \\ 
\textbf{$\pm$0.22} & $\pm$0.20 & $\pm$0.12 & $\pm$0.05 \\
\hline
\multicolumn{4}{c}{\text{50\% of Data on Basel Client:}} \\
\hline
\textbf{0.618} & 0.522 & 0.664 & 0.641 \\ 
\textbf{$\pm$0.23} & $\pm$0.25 & $\pm$0.13 & $\pm$0.08 \\
\hline
\multicolumn{4}{c}{\text{100\% of Data on Basel Client:}} \\
\hline
\textbf{0.608} & 0.365 & 0.675 & 0.628 \\ 
\textbf{$\pm$0.23} & $\pm$0.19 & $\pm$0.10 & $\pm$0.07 \\
\hline
\end{tabular}
}
\end{table}

Based on the experiment comparing a global FL model and locally trained models across different client data quantities with FLAIR + T1 data from Basel, Freiburg, and Strasbourg (see results in Table \ref{tab:segmentation_performance3}), the following observations are made:

\begin{itemize}
    \item Effect of Client Data Quantity on Dice Score: In the global setting, as the client data quantity increases from 20\% to 50\%, there is a slight improvement in the Dice coefficient across all locations (Basel: 0.598 to 0.618, Freiburg: 0.655 to 0.664, Strasbourg: 0.603 to 0.641). In the local setting, there is also an increase in the Dice coefficient with increasing client data quantity, but the values are consistently lower compared to the global (FL) setting for all client sizes (e.g., Basel: 0.475 for 20\%, 0.522 for 50\%, 0.365 for 100\%).
    \item Variability Across Data Quantities: Generally, there is moderate variability in Dice scores across different data quantities, with slightly higher standard deviation values observed in the local model compared to the global model.
\end{itemize}

\textbf{Communication Experiment:}

We briefly present the waiting times experienced by clients, highlighting the duration between the completion of tasks by one client and the initiation of tasks by subsequent clients. The term \textit{Time taken (per 1 round)} refers to the duration it takes for a single client to complete one round of training during the federated learning process. \textit{Waiting time}, on the other hand, refers to the idle time a client experiences while waiting for other clients to finish their training round before the next round can begin.

\textbf{Client-Basel:}
\begin{itemize}
    \item Time taken (per 1 round): $\approx$ 4.08 min
\end{itemize}

\textbf{Client-Freiburg:}
\begin{itemize}
    \item Time taken (per 1 round): $\approx$ 5.47 min
\end{itemize}

\textbf{Client-Strasbourg:}
\begin{itemize}
    \item Time taken (per 1 round): $\approx$ 1.54 min
\end{itemize}

\textit{Waiting Times:}

\textbf{Client-Basel (After Client-Strasbourg):}
\begin{itemize}
    \item Waiting time: $\approx$ 2.54 min
\end{itemize}

\textbf{Client-Freiburg (After Client-Basel):}
\begin{itemize}
    \item Waiting time: $\approx$ 1.39 min
\end{itemize}

In analyzing the performance across Basel, Freiburg, and Strasbourg, notable differences emerge in training times. Client-Strasbourg demonstrates the shortest time, implying potentially superior computational resources. Conversely, Client-Freiburg exhibits the longest training time, hinting at potential resource constraints or computational overhead. Additionally, the waiting times between sequential training rounds highlight the coordination and synchronization efforts inherent in FL setups.

\section{Discussion and Conclusions}

The Clinnova-MS initiative, which promotes cross-border collaboration for personalized MS treatment, faces a unique challenge: utilizing geographically-distributed data while preserving privacy. FL emerges as a powerful tool to address this challenge and unlock the potential of collaborative research within Clinnova-MS.

Under the scope of building POC 1, we focused on establishing the collaborative network between centers and creating the initial virtual infrastructure for further FL experiments. We tested it on the segmentation problem of the prostate in MRIs to assess its functionality and potential issues. In POC 2, we extended this work to real experiments on an MS use case in a geographically distributed setup. Throughout this process, we highlighted the main advantages and challenges we encountered.

\textbf{Advantages of FL for Clinnova-MS:}
\begin{itemize}
\item Enhanced Collaboration: : FL fosters collaboration among multiple healthcare institutions and research centers, facilitating knowledge sharing and joint research efforts. The successful implementation of the POCs for FL within Clinnova-MS lays the groundwork for further collaboration.

\item Continuous Improvement: FL enables continuous learning through iterative model updates based on new data from participating institutions. This allows the collaborative model in Clinnova-MS to continuously adapt and improve over time, reflecting the latest medical advancements and patient data. 

\item Collective Power: FL allows researchers to leverage the collective power of computing resources. This distributed learning approach can significantly accelerate the training of complex models on massive and geographically-dispersed datasets, critical for identifying subtle patterns and developing personalized treatment plans in Clinnova-MS.
    
\item Reduced Data Movement and Bandwidth Constraints: By training models locally, FL minimizes data movement across borders. This not only enhances privacy but also addresses potential bandwidth limitations that could hinder large-scale data transfers in a cross-border initiative like Clinnova-MS.
\end{itemize}

\textbf{Challenges and Limitations: }
Despite its advantages, FL also comes with its set of challenges and limitations.

\begin{itemize}
    \item Initialization: One significant challenge is the initial setup, which can be labor-intensive and requires careful coordination among participating centers. Initiating and coordinating a federation with diverse regulations could be difficult and requires trustworthy communication and compromises. Established contacts between centers prior to building a federation enable building upon an established foundation of trust. 
    
    \item Familiarity with FL:  Additional time and resources may be required to educate the centers about FL concepts and to ensure that they understand the implications and benefits of participating in a federated setup. Having prior familiarity with the concepts, ideally with experienced personnel, is considered advantageous.
    
    \item Data Considerations: Beyond the challenge posed by the sensitivity of the processed data, which mandates robust governance mechanisms to safeguard against unauthorized access and ensure compliance with regulatory requirements (a concern we sidestepped by utilizing a public dataset), there are still other aspects to address. When initiating FL workloads, structured data storage plays a crucial role in facilitating efficient data access. By structuring the data, it becomes easier to locate, retrieve, and manipulate specific data elements needed for FL tasks. Furthermore, structured data storage enhances data consistency and integrity, ensuring that the data used for FL tasks is reliable and accurate.
    
    \item Client Performance and Efficiency: An analysis of client performance and efficiency in FL experiments reveals the importance of optimizing task distribution and coordination. Disparities in computational capabilities among participating clients highlight the need for efficient resource allocation strategies. By minimizing waiting times and maximizing utilization rates, FL can harness the collective computational power of distributed devices to achieve superior model performance.

\end{itemize}

In conclusion, while FL offers huge potential for advancing healthcare research and personalized treatment initiatives, its effective implementation demands a multifaceted approach that addresses both its benefits and challenges comprehensively. By doing so, healthcare organizations can harness the full capabilities of FL while protecting patient privacy and data security, ultimately driving innovation and improving patient outcomes in the Clinnova-MS initiative and beyond.

\section*{Acknowledgements}
This work was partially supported by the Region Grand Est (project CLINNOVA) and by French State Funds managed by the Agence Nationale de la Recherche (ANR) under Grant ANR-10-IAHU-02 (IHU Strasbourg).

We would like to thank the following people for their contributions to the project: Dr. Amandine Bovay, Dr. Thierry Sengstag, Pablo Escobar Lopez, Thanh Nam Bach, Dr. Jean-Luc Dimarcq, and Bastien Andlauer

\bibliography{name}
\bibliographystyle{plain}
\end{document}